\documentclass[runningheads]{llncs}
\usepackage{underscore}

\usepackage[T1]{fontenc}
\usepackage{graphicx}
\usepackage{tikz}
\usetikzlibrary{shapes.geometric, arrows.meta, positioning, fit, backgrounds, calc}
\usepackage{pgfplots}
\pgfplotsset{compat=1.18}
\usepackage{tikz}
\usepackage{colortbl}
\usepackage{xcolor}

\usepackage{color}
\usepackage{url}

\begin{document}
\title{Targeting the Right Situations: Mutation Testing of Fuzzy-Powered Situation Hyperspace}
\titlerunning{Fuzzy-Guided Mutation Testing Framework}
%
\author{Nawshin Mannan Proma\inst{1}\orcidID{0000-0002-8869-3977} }
\authorrunning{N.M. Proma}
%
\institute{University of York, Heslington, United Kingdom
\email{nawshinmannan.proma@york.ac.uk}\\
}
\maketitle              
\begin{abstract}
Ensuring the safety of autonomous Unmanned Aerial Vehicles (UAVs) operating in underground mines is challenging due to the vast and uncertain Operational Design Domain (ODD). Traditional validation approaches, such as random testing, often result in inefficient situation coverage, allocating computational resources to low-risk scenarios while failing to expose critical edge cases. Building on prior work in risk-prioritized Situation Coverage, this paper investigates whether Fuzzy Logic-based risk estimation effectively correlates with real system failures. To this end, we conduct a mutation testing study within our SCALOFT simulation framework. A fixed set of software mutants is systematically injected and evaluated under two testing strategies: (i) uniformly random situation generation and (ii) Fuzzy-Guided risk-prioritized situation generation. Experimental results show that the Fuzzy-Guided approach achieves a higher mutation score and is more effective at exposing difficult-to-detect faults compared to random testing. 


\end{abstract}
\section{Introduction}

The operational deployment of Unmanned Aerial Vehicles (UAVs) in underground mines presents significant challenges for autonomous systems \cite{ryan2024safety}. Ensuring that UAVs behave safely in such conditions therefore requires extensive verification and validation \cite{aslansefat2022}. However, testing the full range of scenarios within the Operational Design Domain (ODD) quickly becomes mathematically and computationally infeasible \cite{SAE_J3016_201806,Khastgir2024_ODD,Elhosary2024_HAZOP}. A common approach in existing work is to rely on random testing, where test scenarios are generated without prioritising particular environmental conditions. Although this method explores many possible situations, it is often inefficient because it treats all states of the environment equally. As a result, testing frequently focuses on relatively safe scenarios, such as wide tunnels with good lighting, where system failures are unlikely, while more critical situations receive less attention. Consequently, large amounts of computational effort may be spent on tests that provide limited insight into the system’s safety.

In our prior work, we addressed this limitation by introducing the Situation Hyperspace \cite{Rob2015,nawshin2023,proma2025scaloft}, which enables systematic exploration of the ODD through the generation of a coverage grid matrix. Building on this, we move towards a systematic, quantitative, and automated approach for risk-based test case selection using fuzzy logic \cite{nawshin_fuzz2026}.This approach models environmental conditions and assigns risk levels to different regions of the situation space, enabling the identification of risk boundaries areas where combinations of environmental factors are more likely to challenge the system’s safety mechanisms. By prioritising these regions, testing can focus on scenarios that are more likely to reveal system vulnerabilities.

However, an important open question remains: does risk-guided scenario selection using fuzzy logic actually improve the effectiveness of safety testing? In this paper, we address this question through empirical validation using mutation testing. Specifically, we introduce software mutants into the drone's control logic within the SCALOFT simulation  framework \cite{proma2025scaloft} and evaluate their impact under high-risk environmental situations generated from a fuzzy-guided hyperspace, comparing the fault-detection capability against traditional random testing.


\section{Mutation Testing Framework}

\begin{figure}
   
\begin{tikzpicture}[
    node distance=0.3cm and 0.5cm, 
    font=\small\sffamily,
    process/.style={
        rectangle, 
        minimum width=2.5cm, 
        minimum height=1cm, 
        text centered, 
        draw=black, 
        fill=orange!10, 
        rounded corners,
        align=center
    },
    data/.style={
        cylinder, 
        shape border rotate=90, 
        aspect=0.25, 
        draw=black, 
        fill=blue!10, 
        text centered, 
        minimum height=1.2cm, 
        minimum width=1.8cm,
        align=center
    },
    decision/.style={
        diamond, 
        aspect=2, 
        draw=black, 
        fill=yellow!10, 
        text centered, 
        minimum width=2.5cm,
        align=center
    },
    engine/.style={
        rectangle, 
        draw=black, 
        fill=green!10, 
        minimum width=5cm, 
        minimum height=1.5cm, 
        text centered, 
        dashed,
        align=center
    },
    arrow/.style={thick, ->, >=Stealth, rounded corners}
]

    \node[data] (odd) {ODD Parameters\\(Light, Width)};
    \node[process, right=of odd] (fuzzy) {Fuzzy Logic\\Engine};
    \node[process, right=of fuzzy] (hyper) {Targeted Situation\\Hyperspace};
    
    \node[decision, below=of hyper] (filter) {Risk Filter\\$R > 0.7$};

    \node[data, left=of odd, xshift=-1cm] (code) {PX4 Source\\Code};
    \node[process, below=of code] (injector) {Mutation\\Engine};
    \node[data, below=of injector] (mutants) {Mutants\\(M1...M10)};

    \node[process, left=of filter, xshift=-0.5cm] (random) {Random\\Generator};
    
    \node[engine, below=of random, yshift=-1cm, xshift=1cm] (scaloft) {\textbf{SCALOFT Framework}\\(Gazebo Simulator + Safety Monitor)};
    
    \node[process, right=of scaloft, xshift=1cm] (result) {Mutation\\Score};

    
    \draw[arrow] (odd) -- (fuzzy);
    \draw[arrow] (fuzzy) -- (hyper);
    \draw[arrow] (hyper) -- (filter);
    
    \draw[arrow] (code) -- (injector);
    \draw[arrow] (injector) -- (mutants);

    \draw[arrow] (filter) -- node[midway, right, font=\footnotesize] {High Risk} (scaloft.40); 
    
    \draw[arrow] (random) -- node[midway, left, font=\footnotesize] {Baseline} (scaloft.140); 
    
    \draw[arrow] (mutants) |- node[midway, below, font=\footnotesize] {Faulty Firmware} (scaloft.west);

    \draw[arrow] (scaloft) -- node[midway, above, font=\footnotesize] {Crash Logs} (result);

\end{tikzpicture}
\caption{Overall Framework for Mutation Testing} \label{framework}
\end{figure}
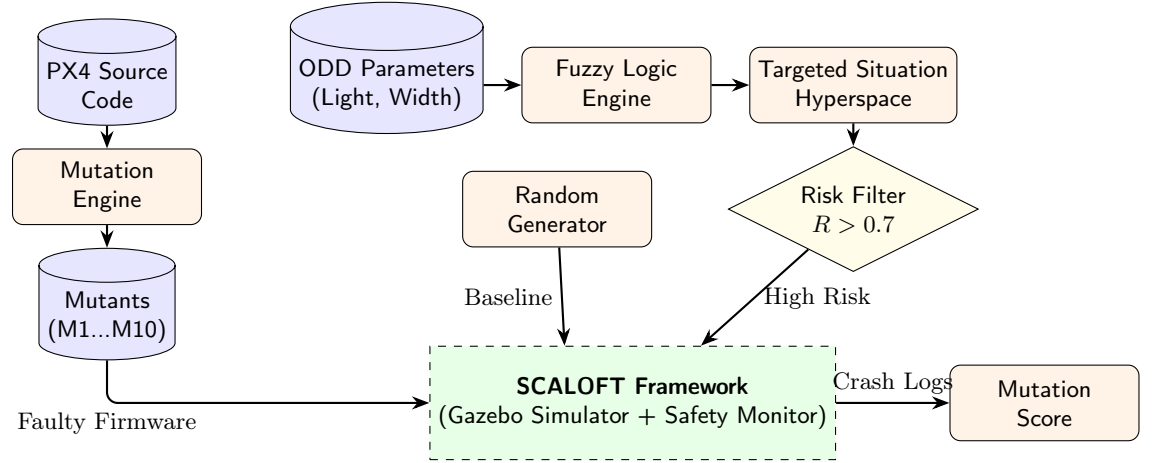

This section describes the methodology used to evaluate the proposed risk-prioritized situation coverage approach. The overall workflow of the framework is illustrated in Fig. \ref{framework}. The framework integrates mutation testing, fuzzy-logic-based risk estimation, and simulation-based testing to focus validation efforts on operational situations that are more likely to expose safety-critical failures in autonomous UAV systems. The process begins with the generation of faulty versions of the UAV control software through mutation testing. Each mutant contains a single injected fault representing a potential implementation error. These faults simulate realistic mistakes that could occur during software development.

Examples of injected faults include logical condition inversions, modifications of arithmetic operators, and changes that increase sensitivity to sensor noise (see table \ref{tab:mutants}). Each mutated version of the controller is compiled into a separate firmware instance and deployed within the SCALOFT simulation environment during testing.

\begin{table}[h]
\centering
\caption{Injected Mutants Used for Evaluation}
\label{tab:mutants}
\begin{tabular}{|c|c|p{5.5cm}|}
\hline
\textbf{Mutant ID} & \textbf{Type} & \textbf{Description} \\ \hline

M1 & AOR & Arithmetic operator modification affecting numerical computations \\ \hline
M2 & BCR & Boundary condition change altering control thresholds \\ \hline
M3 & ASR & State/assignment modification causing abnormal system behaviour \\ \hline
M4 & COD & Conditional logic modification leading to incorrect decisions \\ \hline
M5 & AOR+ASR & Combination of arithmetic and state-related faults \\ \hline

\end{tabular}
\end{table}

A fuzzy logic model is used to estimate the risk level associated with different situations in the ODD. Fuzzy logic is well suited for this task because the relationship between environmental conditions and operational risk is often uncertain and difficult to represent using precise mathematical models.

The fuzzy logic engine takes the ODD parameters as inputs and evaluates them using a set of linguistic rules that represent expert knowledge about hazardous operating conditions. For example, scenarios with very low lighting and narrow tunnels may pose greater risks for UAV navigation than scenarios with sufficient lighting and wider tunnels.

Based on these rules, the fuzzy inference system produces a risk score \(R\) for each situation, indicating the relative likelihood that the given operating condition may lead to unsafe behavior or system failure.
\begin{figure}
    \centering
    \includegraphics[width=0.7\linewidth]{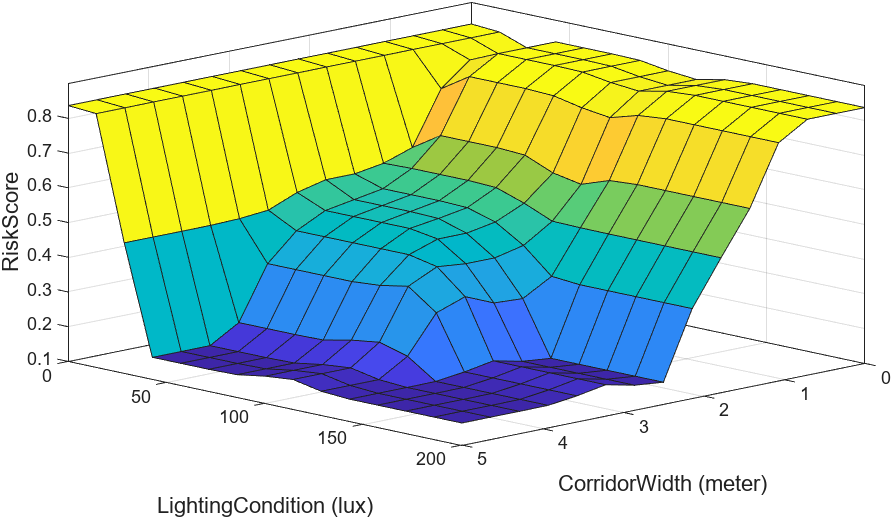} 
    \caption{Fuzzy-powered situation hyperspace from \cite{nawshin_fuzz2026}. The color gradient represents a gradual risk transition (the yellow zone indicates high-risk situations, the cyan region corresponds to medium risk, and the blue region denotes low-risk situations).}
    \label{fig:hyperspace}
\end{figure}

Once the risk score is computed, the framework applies a risk filter to select situations that exceed a predefined threshold. In this study, situations are considered high-risk when

\begin{equation}
R > 0.7
\end{equation}

Only scenarios that satisfy this condition are selected for simulation-based testing. This process forms a \textit{targeted situation hyperspace}, consisting of operational conditions that are most relevant for safety evaluation.

For comparison purposes, a baseline testing strategy is also implemented. In the baseline method, situations are generated randomly across the ODD without considering their associated risk level. All generated scenarios are executed within the \textit{SCALOFT} framework, which integrates a simulation environment and a runtime safety monitoring module. The simulation environment models UAV flight operations in underground mine conditions and executes the mutated versions of the PX4 flight controller.To evaluate the effectiveness of the testing strategies, the framework computes a mutation score that measures the proportion of injected faults detected during testing.
A higher mutation score indicates that the test suite is more effective at exposing faults in the UAV control software. By comparing mutation scores obtained from random testing and fuzzy-guided testing, we evaluate whether prioritizing high-risk situations improves the fault detection capability of the testing process.

\section{Results and Analysis}

The effectiveness of the fuzzy-guided testing approach is evaluated by comparing its mutation score against a random testing baseline. We utilize the Mutation Score ($MS$). Given the stochastic nature of our testing environment, we define $MS$ as the aggregate detection rate across all mutants and test runs:

\begin{equation}
MS = \frac{\sum_{i=1}^{|M|} K_i}{|M| \times |R|} \times 100
\end{equation}

where $|M|$ represents the total number of injected mutants ($M=5$), $|R|$ is the number of independent test runs per mutant ($R=10$), and $K_i$ is the number of successful kills for mutant $i$. Under this metric, the mutation score accounts for both the ability to trigger a fault and the consistency of the testing method in detecting it across multiple scenarios.

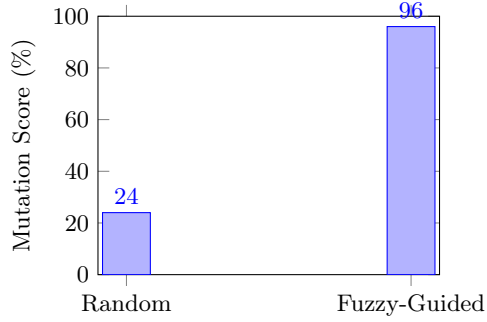
\begin{figure}[t]
\centering
\begin{tikzpicture}
\begin{axis}[
    ybar,
    bar width=18pt,
    width=0.5\linewidth,
    height=5.0cm,
    ylabel={Mutation Score (\%)},
    symbolic x coords={Random, Fuzzy-Guided},
    xtick=data,
    ymin=0, ymax=100,
    nodes near coords,
    nodes near coords align={vertical},
]
\addplot coordinates {(Random,24) (Fuzzy-Guided,96)};
\end{axis}
\end{tikzpicture}
\caption{Comparison of mutation scores between random and fuzzy-guided testing.}
\label{fig:mutation_score}
\end{figure}

As shown in Fig.~\ref{fig:mutation_score}, the fuzzy-guided approach achieves a significantly higher mutation score, successfully killing nearly all injected mutants (M1--M5), whereas random testing detects only a subset of these faults. This difference is further illustrated in Fig.~\ref{fig:mutant_heatmap}, where several mutants remain undetected under random testing but are consistently detected using the risk-guided approach.As illustrated in Fig. \ref{fig:mutation_score}, the fuzzy-guided approach achieves a mutation score of $96\%$, significantly outperforming the random testing baseline of $24\%$ . This score is derived from the aggregate detection frequency across all mutants (M1--M5) over 10 runs each. While random testing often fails to trigger mutants in low-risk scenarios, the fuzzy-guided approach consistently exposes faults by steering the UAV into high-risk regions of the operational design domain.

\begin{figure}[t]
\centering
\begin{tabular}{|c|c|c|}
\hline
\textbf{Mutant} & \textbf{Random (10 Runs)} & \textbf{Fuzzy-Guided (10 Runs)} \\ \hline

M1 & \cellcolor{yellow!40} 4/10 & \cellcolor{green!70} 10/10 \\ \hline
M2 & \cellcolor{red!40} 2/10 & \cellcolor{green!70} 9/10 \\ \hline
M3 & \cellcolor{red!40} 1/10 & \cellcolor{green!70} 10/10 \\ \hline
M4 & \cellcolor{yellow!50} 5/10 & \cellcolor{green!70} 9/10 \\ \hline
M5 & \cellcolor{red!40} 0/10 & \cellcolor{green!70} 10/10 \\ \hline

\end{tabular}
\caption{Mutant detection frequency across generated scenarios. Each cell shows the number of scenarios (out of 10) in which the mutant was killed.}
\label{fig:mutant_heatmap}
\end{figure}

The observed performance gap can be explained by the distribution of generated test scenarios. Random testing samples uniformly across the ODD, resulting in a large proportion of low-risk situations, such as well-lit environments and wide tunnels, where the UAV operates safely. In these conditions, many injected faults are not activated, leading to lower fault detection capability.

In contrast, the fuzzy-guided approach selectively targets high-risk regions of the situation hyperspace. These regions correspond to challenging operational conditions, such as low lighting and narrow tunnels, which place greater stress on the UAV’s control logic and increase the likelihood of exposing latent faults. As a result, faults introduced through mutation are more effectively triggered and detected under fuzzy-guided testing.

Overall, the results demonstrate that fuzzy logic-based risk estimation correlates strongly with failure-prone situations and significantly improves the efficiency and effectiveness of testing compared to uniform random sampling.

\section{Conclusion}

This work bridges the gap between theoretical risk modeling and practical safety testing. We have demonstrated that Fuzzifying the ODD is not merely a theoretical exercise but a practical requirement for efficient validation. The Risk-Prioritized Situation Coverage framework allows safety engineers to validate autonomous mining drones against the most critical scenarios first, ensuring a higher level of safety assurance with reduced computational effort.

%
%
%
%
\bibliographystyle{splncs04}
\bibliography{references}
\end{document}